\documentclass[10pt,twocolumn,letterpaper]{article}

\usepackage{cvpr}
\usepackage{times}
\usepackage{epsfig}
\usepackage{graphicx}
\usepackage{amsmath}
\usepackage{amssymb}
\usepackage{bm}

\usepackage{multirow}
\usepackage{booktabs}
\usepackage{soul}
\usepackage{comment}

\usepackage[pagebackref=true,breaklinks=true,letterpaper=true,colorlinks,bookmarks=false]{hyperref}

\cvprfinalcopy 

\def\I{{\bf{I}}}

\ifcvprfinal\fi
\begin{document}

\title{Deep Reinforcement Learning-based Image Captioning with Embedding Reward}

\author{Zhou Ren$^1$\ \ \ \ \ \ \ \ \ Xiaoyu Wang$^1$\ \ \ \ \ \ \ \ \ Ning Zhang$^1$\ \ \ \ \ \ \ \ \ Xutao Lv$^1$\ \ \ \ \ \ \ \ \ Li-Jia Li$^{2}$\thanks{This work was done when the author was at Snap Inc.} \\
$^1$Snap Inc. \ \ \ \ \ \ \ \ \ \ \ \ \ \ \ $^2$Google Inc.\\
{\tt\small \{zhou.ren, xiaoyu.wang, ning.zhang, xutao.lv\}@snap.com\ \ \ \ \ lijiali@cs.stanford.edu}
}

\maketitle
 
\begin{abstract}
   Image captioning is a challenging problem owing to the complexity in understanding the image content and diverse ways of describing it in natural language. Recent advances in deep neural networks have substantially improved the performance of this task. Most state-of-the-art approaches follow an encoder-decoder framework, which generates captions using a sequential recurrent prediction model. However, in this paper, we introduce a novel decision-making framework for image captioning. We utilize a ``policy network" and a ``value network" to collaboratively generate captions. The policy network serves as a local guidance by providing the confidence of predicting the next word according to the current state. Additionally, the value network serves as a global and lookahead guidance by evaluating all possible extensions of the current state.  
In essence, it adjusts the goal of predicting the correct words towards the goal of generating captions similar to the ground truth captions. We train both networks using an actor-critic reinforcement learning model, with a novel reward defined by visual-semantic embedding. Extensive experiments and analyses on the Microsoft COCO dataset show that the proposed framework outperforms state-of-the-art approaches across different evaluation metrics. 
\end{abstract}

\section{Introduction}
Image captioning, the task of automatically describing the content of an image with natural language, has attracted increasingly interests in computer vision. It is interesting because it aims at endowing machines with one of the core human intelligence to understand the huge amount of visual information and to express it in natural language.

\begin{figure}
  \centering
  \includegraphics[width=0.47\textwidth] {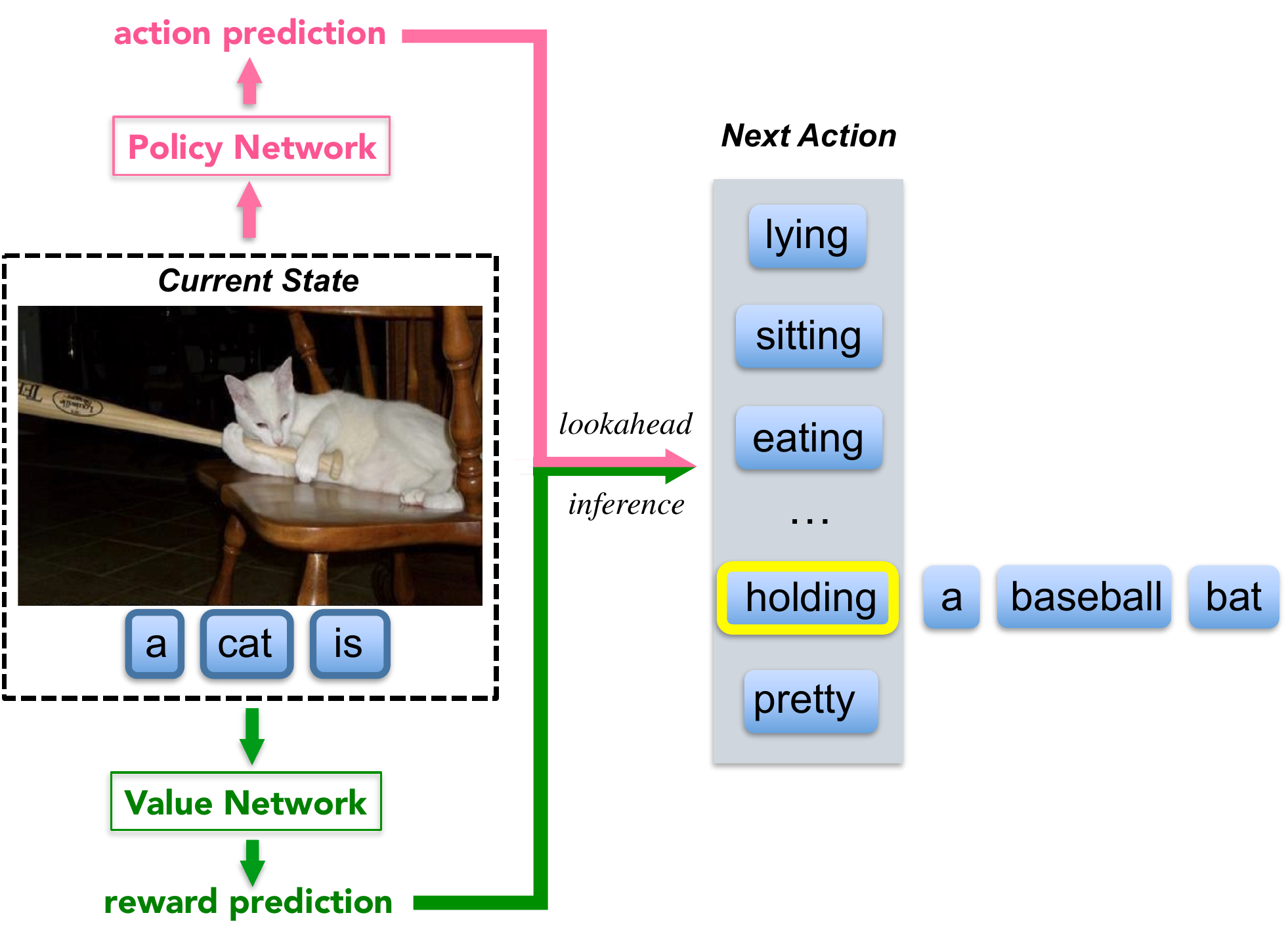}
  \vspace{-2pt}
\caption{ Illustration of the proposed decision-making framework. During our lookahead inference procedure, we utilize a ``policy network" and a ``value network" to collaboratively predict the word for each time step. The policy network provides an action prediction that locally predicts the next word according to current state. The value network provides a reward prediction that globally evaluates all possible extensions of the current state. }
\label{figIntro}
\vspace{-10pt}
\end{figure}

Recent state-of-the-art approaches~\cite{MindEye_CVPR15, GoogleNIC_CVPR15, MRNN_ICLR15, karpathy_cvpr15, LRCN_cvpr15, SpatialAttention_ICML15_Xu, gLSTM_iccv15, SemanticAttention_CVPR16,DBS_2016} follow an encoder-decoder framework to generate captions for the images. They generally employ convolutional neural networks to encode the visual information and utilize recurrent neural networks to decode that information to coherent sentences. 
During training and inference, they try to maximize the probability of the next word based on recurrent hidden state. 

In this paper, we introduce a novel decision-making framework for image captioning. Instead of learning a sequential recurrent model to greedily look for the next correct word, we utilize a ``policy network" and a ``value network" to jointly determine the next best word at each time step.
The policy network, which provides the confidence of predicting the next word according to current state, serves as a \emph{local guidance}. The value network, that evaluates the reward value of all possible extensions of the current state, serves as a \emph{global and lookahead guidance}. Such value network adjusts the goal of predicting the correct words towards the goal of generating captions that are similar to ground truth captions. Our framework is able to include the good words that are with low probability to be drawn by using the policy network alone. Figure~\ref{figIntro} shows an example to illustrate the proposed framework. The word \emph{holding} is not among the top choices of our policy network at current step. But our value network goes forward for one step to the state supposing \emph{holding} is generated and evaluates how good such state is for the goal of generating a good caption in the end. 
The two networks complement each other and are able to choose the word \emph{holding}.

To learn the policy and value networks, we use deep reinforcement learning with embedding reward. We begin by pretraining a policy network using standard supervised learning with cross entropy loss, and by pretraining a value network with mean squared loss. Then, we improve the policy and value networks by deep reinforcement learning. Reinforcement learning has been widely used in gaming~\cite{AlphaGo_nature16}, control theory~\cite{Minh_nature15_gamingDRL}, \emph{etc}. The problems in control or gaming have concrete targets to optimize by nature, whereas defining an appropriate optimization goal is nontrivial for image captioning. In this paper, we propose to train using an actor-critic model~\cite{actor_critic_99} with reward driven by visual-semantic embedding~\cite{DeviSE, unifying_embedding_acl15, MIVSE_15, GVSE_mm16}. Visual-semantic embedding, which provides a measure of similarity between images and sentences, can measure the correctness of generated captions and serve as a reasonable global target to optimize for image captioning in reinforcement learning.

We conduct detailed analyses on our framework to understand its merits and properties. Extensive experiments on the Microsoft COCO dataset~\cite{mscoco_eccv14} show that the proposed method outperforms state-of-the-art approaches consistently across different evaluation metrics, including BLEU \cite{bleu_acl02}, Meteor~\cite{meteor_10}, Rouge~\cite{rouge_04} and CIDEr~\cite{cider_cvpr15}. The contributions of this paper are summarized as follows: 
\begin{itemize}
\vspace{-5pt}
 \item We present a novel decision-making framework for image captioning utilizing a policy network and a value network. Our method achieves state-of-the-art performance on the MS COCO dataset. To our best knowledge, this is the first work that applies decision-making framework to image captioning. 
\vspace{-5pt}
 \item To learn our policy and value networks, we introduce an actor-critic reinforcement learning algorithm driven by visual-semantic embedding. Our experiments suggest that the supervision from embedding generalizes well across different evaluation metrics. 
 \end{itemize} 
 \vspace{-10pt}
 

\section{Related Work}
 \vspace{-2pt}
\subsection {Image captioning}
 \vspace{-2pt}

Many image captioning approaches have been proposed in the literature. Early approaches tackled this problem using bottom-up paradigm~\cite{PicStory_ECCV10, BabyTalk_cvpr11, ngram_conll11, Yang_emnlp11, collectiveGen_acl12, elliott_emnlp13, phraseCaption_iclr15, caption2concept_cvpr15}, which first generated descriptive words of an image by object recognition and attribute prediction, and then combined them by language models. 
Recently, inspired by the successful use of neural networks in machine translation~\cite{RNN_machineTransl_emnlp14}, the encoder-decoder framework~\cite{MindEye_CVPR15, GoogleNIC_CVPR15, MRNN_ICLR15, karpathy_cvpr15, LRCN_cvpr15, SpatialAttention_ICML15_Xu, gLSTM_iccv15, SemanticAttention_CVPR16,DBS_2016} has been brought to image captioning. Researchers adopted such framework because ``translating" an image to a sentence was analogous to the task in machine translation. Approaches following this framework generally encoded an image as a single feature vector by convolutional neural networks~\cite{Alexnet12, Imagenet09, vggnet15, googlenet15}, and then fed such vector into recurrent neural networks~\cite{LSTM_97,GRU_2014} to generate captions. On top of it, various modeling strategies have been developed. Karpathy and Fei-Fei~\cite{karpathy_cvpr15}, Fang~\emph{et al.}~\cite{caption2concept_cvpr15} presented methods to enhance their models by detecting objects in images. To mimic the visual system of humans~\cite{VisualAttention_Koch87}, spatial attention~\cite{SpatialAttention_ICML15_Xu} and semantic attention~\cite{SemanticAttention_CVPR16} were proposed to direct the model to attend to the meaningful fine details. Dense captioning~\cite{denseCap_cvpr16} was proposed to handle the localization and captioning tasks simultaneously.
Ranzato~\emph{et al.}~\cite{MIXER_ICLR16} proposed a sequence-level training algorithm.

During inference, most state-of-the-art methods employ a common decoder mechanism using greedy search or beam search. Words are sequentially drawn according to local confidence. Since they always predict the words with top local confidence, such mechanism can miss good words at early steps which may lead to bad captions. In contrast, in addition to the local guidance, our method also utilizes a global and lookahead guidance to compensate such errors. 
 \vspace{-3pt}
 
\subsection {Decision-making}
 \vspace{-3pt}
 
Decision-making is the core problem in computer gaming~\cite{AlphaGo_nature16}, control theory~\cite{Minh_nature15_gamingDRL}, navigation and path planning~\cite{YukeZhu16_VisualNavigation}, \emph{etc}. In those problems, there exist agents that interact with the environment, execute a series of actions, and aim to fulfill some pre-defined goals. Reinforcement learning~\cite{Reinforce_92,actor_critic_99,PolicyGradientMethod_DRL_nips00,a3c_icml16}, known as ``a machine learning technique concerning how software agent ought to take actions in an environment so as to maximize some notion of cumulative reward", is well suited for the task of decision-making. Recently, professional-level computer Go program was designed by Silver~\emph{et al.}~\cite{AlphaGo_nature16} using deep neural networks and Monte Carlo Tree Search. Human-level gaming control~\cite{Minh_nature15_gamingDRL} was achieved through deep Q-learning. A visual navigation system~\cite{YukeZhu16_VisualNavigation} was proposed recently based on actor-critic reinforcement learning model. 


Decision-making framework has not been applied to image captioning. One previous work in text generation~\cite{MIXER_ICLR16} has used REINFORCE~\cite{Reinforce_92} to train its model by directly optimizing a user-specified evaluation metric. Such metric-driven approach~\cite{MIXER_ICLR16} is hard to generalize to other metrics. To perform well across different metrics, it needs to be re-trained for each one in isolation. In this paper, we propose a training method using actor-critic reinforcement learning~\cite{actor_critic_99} driven by visual-semantic embedding~\cite{DeviSE, unifying_embedding_acl15}, which performs well across different evaluation metrics without re-training. Our approach shows significant performance improvement over \cite{MIXER_ICLR16}. Moreover, we use a decision-making framework to generate captions, while \cite{MIXER_ICLR16} follows the existing encoder-decoder framework.

\section{Deep Reinforcement Learning-based Image Captioning}
In this section, we first define our formulation for deep reinforcement learning-based image captioning and propose a novel reward function defined by visual-semantic embedding. Then we introduce our training procedure as well as our inference mechanism.

\subsection{Problem formulation}
We formulate image captioning as a decision-making process. In decision-making, there is an \emph{agent} that interacts with the \emph{environment}, and executes a series of \emph{actions}, so as to optimize a \emph{goal}. In image captioning, the goal is, given an image $\I$, to generate a sentence $S=\{w_1, w_2, ..., w_T\}$ which correctly describes the image content, where $w_i$ is a word in sentence $S$ and $T$ is the length. Our model, including the policy network $p_\pi$ and value network $v_\theta$, can be viewed as the agent; the environment is the given image $\I$ and the words predicted so far $\{w_1,...,w_t\}$; and an action is to predict the next word $w_{t+1}$.
\vspace{-5pt}

\begin{figure}
  \centering
  \includegraphics[width=0.43\textwidth] {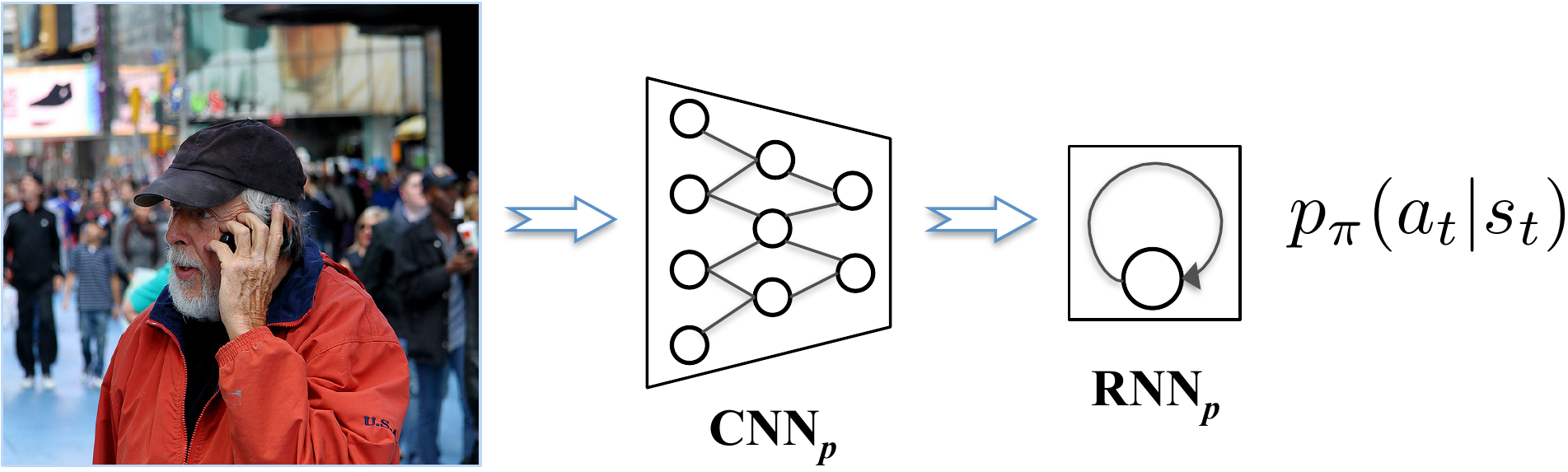}
\caption{An illustration of our policy network $p_\pi$ that is comprised of a CNN and a RNN. The CNN$_p$ output is fed as the initial input of RNN$_p$. The policy network computes the probability of executing an action $a_t$ at a certain state $s_t$, by $p_\pi(a_t|s_t)$.}
\label{FigPN}
\vspace{-0pt}
\end{figure}

\subsubsection{State and action space}
A decision-making process consists of a series of actions. After each action $a$, a state $s$ is observed. In our problem, state $s_t$ at time step $t$ consists of the image $\I$ and the words predicted until $t$, $\{w_1,...,w_t\}$. The action space is the dictionary $\mathcal{Y}$ that the words are drawn from, \emph{i.e.}, $a_t\subset\mathcal{Y}$.
\vspace{-4pt}

\begin{figure}
  \centering
  \includegraphics[width=0.43\textwidth] {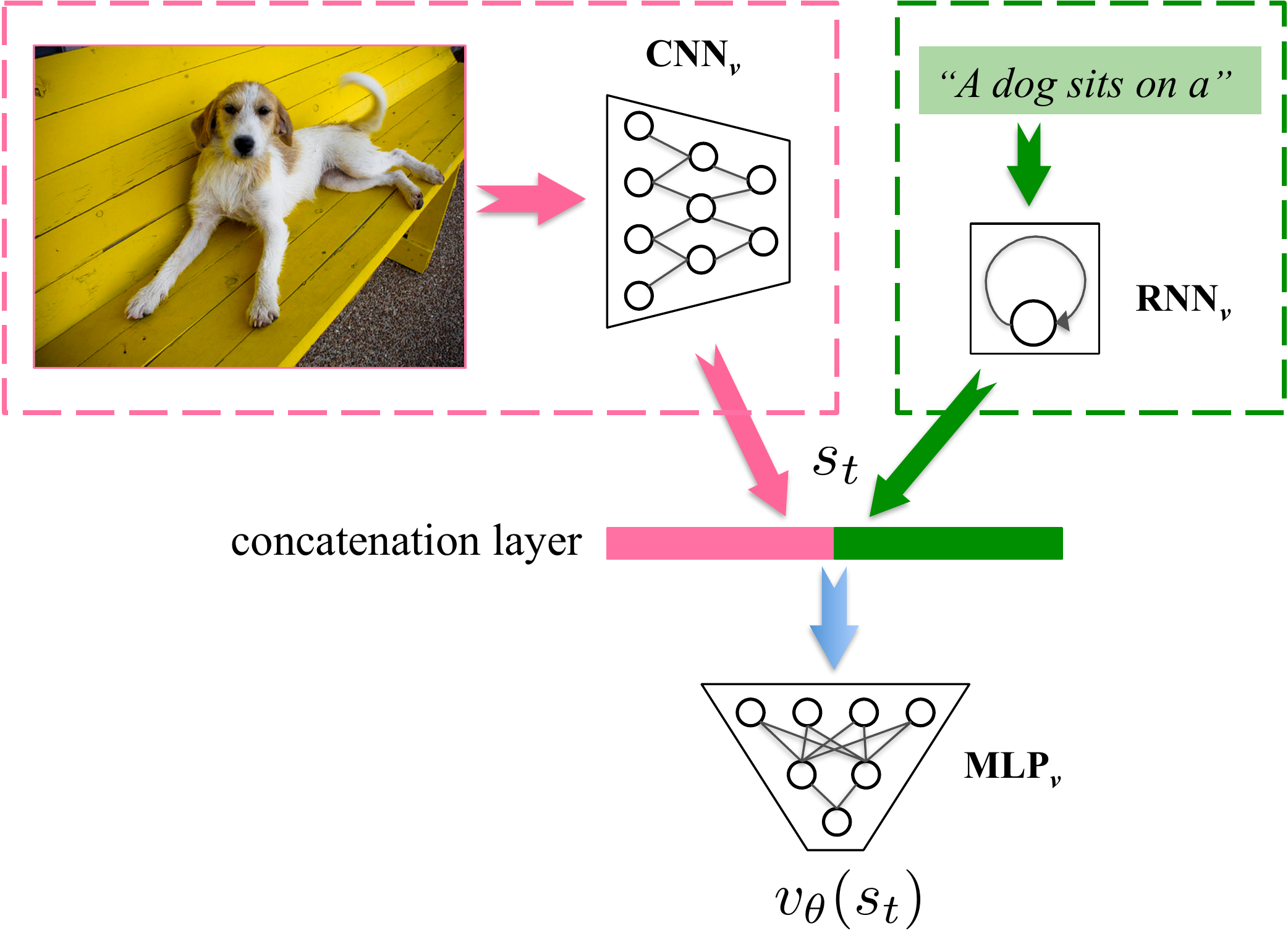}
  \vspace{-3pt}
\caption{ An illustration of our value network $v_\theta$ that is comprised of a CNN, a RNN and a MLP. Given a state $s_t$ which contains raw image input $\I$ and a partially generated raw sentence until $t$, the value network $v_\theta(s_t)$ evaluates its value.}
\label{FigVN}
\vspace{-5pt}
\end{figure}

\subsubsection{Policy network}
The policy network $p_\pi$ provides the probability for the agent to take actions at each state, $p_\pi(a_t|s_t)$, where the current state $s_t=\{\I,w_1,...,w_t\}$ and action $a_t=w_{t+1}$. In this paper, we use a Convolutional Neural Network (CNN) and a Recurrent Neural Network (RNN) to construct our policy network, denoted as $\text{CNN}_p$ and RNN$_p$. It is similar to the basic image captioning model \cite{GoogleNIC_CVPR15} used in the encoder-decoder framework. As shown in Figure~\ref{FigPN}, firstly we use CNN$_p$ to encode the visual information of image $\I$. The visual information is then fed into the initial input node $\bm{x}_0\in\mathbb{R}^n$ of RNN$_p$. As the hidden state $\bm{h}_t\in\mathbb{R}^m$ of RNN$_p$ evolves over time $t$, the policy at each time step to take an action $a_t$ is provided. The generated word $w_t$ at $t$ will be fed back into RNN$_p$ in the next time step as the network input $\bm{x}_{t+1}$, which drives the RNN$_p$ state transition from $\bm{h}_t$ to $\bm{h}_{t+1}$. Specifically, the main working flow of $p_{\pi}$ is governed by the following equations: 
\begin{align}
 &\bm{x}_0 = \bm{W}^{x,v}\text{CNN}{_p}(\I) \label{a}\\
 &\bm{h}_t  = \textnormal{RNN}{_p}(\bm{h}_{t-1}, \bm{x}_t) \label{b}\\
 &\bm{x}_t = \phi(w_{t-1}),\ \ t>0 \label{c}\\
 &p_\pi(a_t|s_t) =  \varphi(\bm{h}_t) \label{d}
 \end{align}
where $ \bm{W}^{x,v}$ is the weight of the linear embedding model of visual information, $\phi$ and $\varphi$ denote the input and output models of RNN$_p$.
\vspace{-4pt}

\subsubsection{Value network}
Before we introduce our value network $v_\theta$, we first define the value function $v^p$ of a policy $p$. $v^p$ is defined as the prediction of the total reward $r$ (will be defined later in Section~\ref{reward}) from the observed state $s_t$, assuming the decision-making process is following a policy $p$, \emph{i.e.},
\begin{equation}
v^p(s)= \mathbb{E}[r|s_t=s,\ \ a_{t...T}\sim p]
\end{equation}

We approximate the value function using a value network, $v_\theta(s)\approx v^p(s)$. It serves as an evaluation of state $s_t=\{\I,w_1,...,w_t\}$. As shown in Figure~\ref{FigVN}, our value network is comprised of a CNN, a RNN, and a Multilayer Perceptron (MLP), denoted as CNN$_v$, RNN$_v$ and MLP$_v$. Our value network takes the raw image and sentence inputs. CNN$_v$ is utilized to encode the visual information of $\I$, RNN$_v$ is designed to encode the semantic information of a partially generated sentence $\{w_1,...,w_t\}$. All the components are trained simultaneously to regress the scalar reward from $s_t$. We investigate our value network architecture in Section~\ref{netAnalysis}.

\subsection{Reward defined by visual-semantic embedding}
\label{reward}
In our decision-making framework, it is important to define a concrete and reasonable optimization goal, \emph{i.e.,} the \emph{reward} for reinforcement learning. We propose to utilize visual-semantic embedding similarities as the reward.

Visual-semantic embedding has been successfully applied to image classification~\cite{DeviSE, GVSE_mm16}, retrieval~\cite{unifying_embedding_acl15, MIVSE_15, EmbeddingVideo_CVPR16}, \emph{etc}. Our embedding model is comprised of a CNN, a RNN and a linear mapping layer, denoted as ${\text{CNN}_e}$, ${\text{RNN}_e}$ and $f_e$. By learning the mapping of images and sentences into one semantic embedding space, it provides a measure of similarity between images and sentences. Given a sentence $S$, its embedding feature is represented using the last hidden state of {$\text{RNN}_e$}, \emph{i.e.}, $\bm{h}^\prime_T(S)$. Let $\bm{v}$ denote the feature vector of image $\I$ extracted by $\text{CNN}_e$, and $f_e(\cdot)$ is the mapping function from image features to the embedding space. We train the embedding model using the same image-sentence pairs as in image captioning. We fix the CNN$_e$ weight, and learn the RNN$_e$ weights as well as $f_e(\cdot)$ using a bi-directional ranking loss defined as follows: 
\vspace{-9pt}

\begin{small}
\begin{align}
\!\!\!\!L_e\!=\!&\sum_{\bm{v}}\sum_{S^-} \max(0,\ \beta\!-\!f_e(\bm{v})\cdot\bm{h}^\prime_T(S)\!+\! f_e(\bm{v})\cdot\bm{h}^\prime_T(S^-)) \nonumber \\
+\!& \sum_{S}\sum_{\bm{v}^-} \max(0,\ \beta\!-\!\bm{h}^\prime_T(S)\cdot f_e(\bm{v})\!+\!\bm{h}^\prime_T(S)\cdot f_e(\bm{v}^-)) 
\label{Bi_directionLossFunc}
\end{align}
\end{small}
\!\!\!where $\beta$ is the margin cross-validated, every ($\bm{v}, S$) are a ground truth image-sentence pair, $S^-$ denotes a negative description for the image corresponding to $\bm{v}$, and vice-versa with $\bm{v}^-$.

Given an image with feature $\bm{v}^*$, we define the reward of a generated sentence $\widehat{S}$ to be the embedding similarity between $\widehat{S}$ and ${\bm{v}^*}$: 
\vspace{-5pt}
\begin{align}
r= \frac{f_e({\bm{v}^*})\cdot \bm{h}^\prime_T(\widehat{S})}{\|f_e({\bm{v}^*})\| \|\bm{h}^\prime_T(\widehat{S})\|}
\end{align}

\subsection{Training using deep reinforcement learning}
\label{training}
Following~\cite{AlphaGo_nature16}, we learn $p_\pi$ and $v_\theta$ in two steps. In the first step, we train the policy network $p_\pi$ using standard supervised learning with cross entropy loss, where the loss function is defined as $L_{p^\prime}=-\text{log}\ p(w_1,...,w_T|\I; \pi)=-\sum_{t=1}^T\text{log}\ p_\pi(a_t|s_t)$. And we train the value network by minimizing the mean squared loss, $||v_\theta(s_i)-r||^2$ where $r$ is the final reward of the generated sentence and $s_i$ denotes a \emph{randomly selected} state in the generating process. For one generated sentence, successive states are strongly correlated, differing by just one word, but the regression target is shared for each entire captioning process. Thus, we randomly sample one single state from each distinct sentence, to prevent overfitting.

In the second step, we train $p_\pi$ and $v_\theta$ jointly using deep reinforcement learning (RL). The parameters of our agent are represented by $\Theta=\{\pi,\theta\}$, and we learn $\Theta$ by maximizing the total reward the agent can expect when interacting with the environment: $J(\Theta)=\mathbb{E}_{s_{1...T}\sim p_\pi}(\sum_{t=1}^Tr_t)$. As $r_t=0\ \ \forall\ 0<t<T\ \text{and}\ r_T=r$, $J(\Theta)=\mathbb{E}_{s_{1...T}\sim p_\pi}(r)$.

Maximizing $J$ exactly is non-trivial since it involves an expectation over the high-dimensional interaction sequences which may involve unknown environment dynamics in turn. Viewing the problem as a partially observable Markov decision process, however, allows us to bring techniques from the RL literature to bear: As shown in~\cite{Reinforce_92, PolicyGradientMethod_DRL_nips00, a3c_icml16}, a sample approximation to the gradient is:
\vspace{-3pt}
\begin{align}
&\nabla_\pi J \approx \sum_{t=1}^T \nabla_\pi \text{log}\ p_\pi(a_t|s_t)\ (r-v_\theta(s_t)) \\
&\nabla_\theta J = \nabla_\theta v_\theta(s_t)\ (r-v_\theta(s_t))
\end{align}

Here the value network $v_\theta$ serves as a moving baseline. The subtraction with the evaluation of value network leads to a much lower variance estimate of the policy gradient. The quantity $r-v_\theta(s_t)$ used to scale the gradient can be seen as an estimate of the advantage of action $a_t$ in state $s_t$. This approach can be viewed as an actor-critic architecture where the policy $p_\pi$ is the actor and $v_\theta$ is the critic.

However, reinforcement learning in image captioning is hard to train, because of the large action space comparing to other decision-making problems. The action space of image captioning is in the order of $10^3$ which equals the vocabulary size, while that of visual navigation in~\cite{YukeZhu16_VisualNavigation} is only 4, which indicates four directions to go. To handle this problem, we follow~\cite{MIXER_ICLR16} to apply curriculum learning~\cite{curriculearning_icml09} to train our actor-critic model. In order to gradually teach the model to produce stable sentences, we provide training samples with gradually more difficulty: iteratively we fix the first ($T-i\times\Delta$) words with cross entropy loss and let the actor-critic model train with the remaining $i\times\Delta$ words, for $i=1,2,...,$ until reinforcement learning is used to train the whole sentence. Please refer to~\cite{MIXER_ICLR16} for details.

\subsection{Lookahead inference with policy network and value network}
One of the key contributions of the proposed decision-making framework over existing framework lies in the inference mechanism. For decision-making problems, the inference is guided by a local guidance and a global guidance, \emph{e.g.}, AlphaGo~\cite{AlphaGo_nature16} utilized MCTS to combine both guidances. 
For image captioning, we propose a novel lookahead inference mechanism that combines the local guidance of policy network and the global guidance of value network. The learned value network provides a lookahead evaluation for each decision, which can complement the policy network and collaboratively generate captions. 

Beam Search (BS) is the most prevalent method for decoding in existing image captioning approaches, which stores the top-$B$ highly scoring candidates at each time step. Here $B$ is the beam width. Let us denote the set of $B$ sequences held by BS at time $t$ as $W_{\lceil t\rceil}=\{\bm{w}_{1,{\lceil t\rceil}}, ..., \bm{w}_{B, {\lceil t\rceil}}\}$, where each sequence are the generated words until then, $\bm{w}_{b,{\lceil t \rceil}}=\{w_{b,1},...,w_{b,t}\}$.  At each time step $t$, BS considers all possible single word extensions of these beams, given by the set $\mathcal{W}_{t+1}=W_{\lceil t\rceil}\times\mathcal{Y}$, and selects the top-$B$ most scoring extensions as the new beam sequences $W_{\lceil t+1\rceil}$: 
\begin{align}
W_{\lceil t+1\rceil}\! =\!\!\!\!  \operatornamewithlimits{arg\emph{topB}}_{\bm{w}_{b,{\lceil t+1 \rceil}}\in \mathcal{W}_{t+1}}\! \!\! S(\bm{w}_{b,{\lceil t+1 \rceil}}), \ \ s.t.\ \bm{w}_{i,{\lceil t+1 \rceil}}\!\neq\! \bm{w}_{j,{\lceil t+1 \rceil}} \nonumber
\end{align}
where operator arg\emph{topB} denotes the obtaining top-$B$ operation that is implemented by sorting the $B\times |\mathcal{Y}|$ members of $\mathcal{W}_{t+1}$, and $S(\cdot)$ denotes the scoring function of a generated sequence. In existing BS of image captioning, $S(\cdot)$ is the log-probability of the generated sequence. However, such scoring function may miss good captions because it assumes that the log-probability of every word in a good caption must be among top choices. This is not necessarily true. Analogously, in AlphaGo not every move is with top probability. It is beneficial to sometimes allow some actions with low probability to be selected as long as the final reward is optimized. 

To this end, we propose a lookahead inference that combines the policy network and value network to consider all options in $\mathcal{W}_{t+1}$. It executes each action by taking both the current policy and the lookahead reward evaluation into consideration, \emph{i.e.}, 
\begin{align}
&S(\bm{w}_{b,{\lceil t+1 \rceil}}) =S(\{\bm{w}_{b,{\lceil t \rceil}}, w_{b,t+1}\}) \label{lookaheadBS} \\ 
&=S(\bm{w}_{b,{\lceil t \rceil}}) +\lambda\ \text{log}\ p_\pi(a_t|s_t)+(1-\lambda)\ v_\theta(\{s_t, w_{b,t+1}\}) \nonumber
\end{align}
where $S(\bm{w}_{b,{\lceil t+1 \rceil}})$ is the score of extending the current sequence $\bm{w}_{b,{\lceil t \rceil}}$ with a word $w_{b,t+1}$, $\text{log}\ p_\pi(a_t|s_t)$ denotes the confidence of policy network to predict $w_{b,t+1}$ as extension, and $v_\theta(\{s_t, w_{b,t+1}\})$ denotes the evaluation of value network for the state supposing $w_{b,t+1}$ is generated. $0\leq\lambda\leq1$ is a hyperparameter combining policy and value network that we will analyze experimentally in Section~\ref{parAnalysis}.

\begin{table*}
\centering
  \small
\begin{tabular}{| l | c c c c c c c | }
\hline
Methods &  Bleu-1 & Bleu-2 & Bleu-3 & Bleu-4 & METEOR & Rouge-L & CIDEr  \\
\hline
Google NIC~\cite{GoogleNIC_CVPR15} & 0.666  & 0.461 & 0.329 & 0.246 & $-$ & $-$ & $-$ \\
 m-RNN~\cite{MRNN_ICLR15} &  0.67 & 0.49 & 0.35 & 0.25 & $-$ & $-$ & $-$  \\
 BRNN~\cite{karpathy_cvpr15} &  0.642 & 0.451 & 0.304 & 0.203 & $-$ & $-$ & $-$  \\
 LRCN~\cite{LRCN_cvpr15} & 0.628 & 0.442 & 0.304 & 0.21 & $-$ & $-$ & $-$  \\
 MSR/CMU~\cite{MindEye_CVPR15} & $-$ & $-$ & $-$ & 0.19 & 0.204 & $-$ & $-$  \\
 Spatial ATT~\cite{SpatialAttention_ICML15_Xu} & \bf{0.718} & 0.504 & 0.357 & 0.25 & 0.23 & $-$ &  $-$ \\
 gLSTM~\cite{gLSTM_iccv15} &0.67 & 0.491 & 0.358 & 0.264 &  0.227 &  $-$ & 0.813\\
 MIXER~\cite{MIXER_ICLR16} & $-$ & $-$ & $-$ & 0.29 & $-$ & $-$ & $-$  \\
 \hline
 Semantic ATT~\cite{SemanticAttention_CVPR16} $^*$& 0.709 & 0.537 & 0.402 & \bf{0.304} & 0.243 & $-$ & $-$ \\
 DCC~\cite{DCC_captioningCVPR2016} $^*$ &0.644 & $-$ & $-$ & $-$ & 0.21 & $-$ & $-$ \\
 \hline
 Ours & 0.713 & \bf{0.539} & \bf{0.403} & \bf{0.304} & \bf{0.251} & \bf{0.525} & \bf{0.937}  \\
\hline
\end{tabular}
\vspace{1pt}
\caption{Performance of our method on MS COCO dataset, comparing with state-of-the-art methods. Our beam size is set to 10. For those competing methods, we show the results from their latest version of paper. The numbers in bold face are the best known results and ($-$) indicates unknown scores. ($^*$) indicates that external data was used for training in these methods.}
\label{table_compareSOTA}
\vspace{0pt}
\end{table*}

\begin{table*}
\centering
 \small
\begin{tabular}{| l | c c c c c c c | }
\hline
Methods &  Bleu{-}1 & Bleu{-}2 & Bleu{-}3 & Bleu{-}4 & METEOR & Rouge{-}L & CIDEr  \\
\hline
SL & 0.692 & 0.519 & 0.384 & 0.289 & 0.237 & 0.512 & 0.872  \\
SL-Embed & 0.7 & 0.523 & 0.383 & 0.280 & 0.241 & 0.514 & 0.888 \\
SL-RawVN & 0.706 & 0.533 & 0.395 & 0.298 & 0.243 & 0.52 & 0.916 \\
 \hline
 hid-VN & 0.603 & 0.429 & 0.292 & 0.197 & 0.2 & 0.467 & 0.69 \\
 hid-Im-VN & 0.611 & 0.435 & 0.297 & 0.201 & 0.202 & 0.468 & 0.701 \\
 \hline
  Full-model & 0.713 & {0.539} & {0.403} & {0.304} & {0.251} & {0.525} & {0.937}  \\
\hline
\end{tabular}
\vspace{1pt}
\caption{Performance of the variants of our method on MS COCO dataset, with beam size = 10. {\bf SL}: supervised learning baseline. {\bf SL-Embed}: {\bf SL} with embedding. {\bf SL-RawVN}: {\bf SL} plus pretrained raw value network.  {\bf hid-VN}: value network directly utilizes policy hidden state. {\bf hid-Im-VN}: value network utilizes policy hidden state and policy image feature. {\bf Full-model}: our full model. }
\label{table_compareVariants}
\vspace{-3pt}
\end{table*}

\section{Experiments}

In this section, we perform extensive experiments to evaluate the proposed framework. All the reported results are computed using Microsoft COCO caption evaluation tool \cite{MSCOCO_server}, including the metrics BLEU, Meteor, Rouge-L and CIDEr, which are commonly used together for fair and thorough performance measure. Firstly, we discuss the dataset and implementation details. Then we compare the proposed method with state-of-the-art approaches on image captioning. Finally, we conduct detailed analyses on our method.

\subsection{Dataset and implementation details}
{{\bf Dataset}} We evaluate our method on the widely used MS COCO dataset \cite{mscoco_eccv14} for {the} image captioning task. For fair comparison, we adopt the {commonly} used splits proposed in ~\cite{karpathy_cvpr15}, which use 82,783 images for training, 5,000 images for validation, and 5,000 images for testing. Each image is given at least five captions by different AMT workers. We follow~\cite{karpathy_cvpr15} to preprocess the captions (\emph{i.e.} building dictionaries, tokenizing the captions).

{\bf{Network architecture}} As shown in Figure~\ref{FigPN} and~\ref{FigVN}, our policy network, value network both contain a CNN and a RNN. We adopt the same CNN and RNN architectures for them, but train them independently. We use VGG-16~\cite{vggnet15} as our CNN architecture and LSTM~\cite{LSTM_97} as our RNN architecture. The input node dimension and the hidden state dimension of LSTM are both set to be 512, \emph{i.e.}, $m=n=512$. There are many CNN, RNN architectures in the literature, \emph{e.g.}, ResNet~\cite{ResNet_cvpr16}, GRU~\cite{GRU_2014}, \emph{etc}. Some of them have reported better performance than the ones we use. We do not use the latest architecture for fair comparison with existing methods. In our value network, we use a three-layer MLP that regresses to {a} scalar reward value, with a 1024-dim and a 512-dim hidden layers in between. In Figure~\ref{FigVN}, a state $s_t$ is represented by concatenating the visual and semantic features. The visual feature is a 512-dim embedded feature, mapped from the 4096-dim CNN$_v$ output. The semantic feature is the 512-dim hidden state of RNN$_v$ at the last time step. Thus, the dimension of $s_t$ is 1024. 

{\bf{Visual-semantic embedding}} {Visual-semantic embedding can measure the similarity between images and sentences by mapping them to the same space.} We followed \cite{unifying_embedding_acl15} to use VGG-16~\cite{vggnet15} as ${\text{CNN}_e}$ and GRU~\cite{GRU_2014} as ${\text{RNN}_e}$. The image feature $\bm{v}$ in Equation~\ref{Bi_directionLossFunc} is extracted from the last 4096-dim layer of VGG-16. The input node dimension and the hidden state dimension of GRU are set as 300 and 1024. $f_e(\cdot)$ is a 4096$\times$1024 linear mapping layer. The margin $\beta$ in Equation~\ref{Bi_directionLossFunc} is set as 0.2.

{\bf{Training details}} In training, we use Adam~\cite{adam_iclr15} algorithm to do model updating. It is worth noting that, other than using the pretrained VGG-16 model, we only use the images and captions provided in the dataset to train our networks and embedding, without any external data. We set $\Delta$ in curriculum learning as 2. In testing, a caption is formed by drawing words sequentially until a special end token is reached, using the proposed lookahead inference mechanism. We do not use ensemble of models.

\subsection{Comparing with state-of-the-art methods}
In Table~\ref{table_compareSOTA}, we provide a summary of the results of our method and existing methods. We obtain state-of-the-art performance on MS COCO in most evaluation metrics. Note that Semantic ATT~\cite{SemanticAttention_CVPR16} utilized rich extra data from social media to train their visual attribute predictor, and DCC \cite{DCC_captioningCVPR2016} utilized external data to prove its unique transfer capacity. It makes their results incomparable to other methods that do not use extra training data. 
Surprisingly, even without external training data, our method outperforms \cite{SemanticAttention_CVPR16,DCC_captioningCVPR2016}. Comparing to methods other than \cite{SemanticAttention_CVPR16,DCC_captioningCVPR2016}, our approach shows significant improvements in all the metrics except Bleu-1 in which our method ranks the second. Bleu-1 is related to single word accuracy, the performance gap of Bleu-1 between our method and \cite{SpatialAttention_ICML15_Xu} may be due to different preprocessing for word vocabularies. MIXER \cite{MIXER_ICLR16} is a metric-driven trained method. A model trained with Bleu-4 using \cite{MIXER_ICLR16} is hard to generalize to other metrics. Our embedding-driven decision-making approach performs well in all metrics. Especially, considering our policy network shown in Figure~\ref{FigPN} is based on a mechanism similar to the very basic image captioning model similar to Google NIC~\cite{GoogleNIC_CVPR15}, such significant improvement over \cite{GoogleNIC_CVPR15} validates the effectiveness of the proposed decision-making framework that utilizes both policy and value networks. Moreover, the proposed framework is modular w.r.t.~the network design. Other powerful mechanisms such as spatial attention, semantic attention can be directly integrated into our policy network and further improve our performance.

Since the proposed embedding-driven decision-making framework is very different from existing methods, we want to perform insightful analyses and answer the following questions: 1) How powerful is embedding? Is the performance gain more because of the framework or embedding alone? 2) How important is lookahead inference? 3) How important is reinforcement learning in the framework? 4) Why the value network is designed as in Figure~\ref{FigVN}? 5) How sensitive is the method to hyperparameter $\lambda$ and beam size? 
To answer those questions, we conduct detailed analyses in the following three sections.

\begin{figure*}
  \centering
  \includegraphics[width=0.96\textwidth] {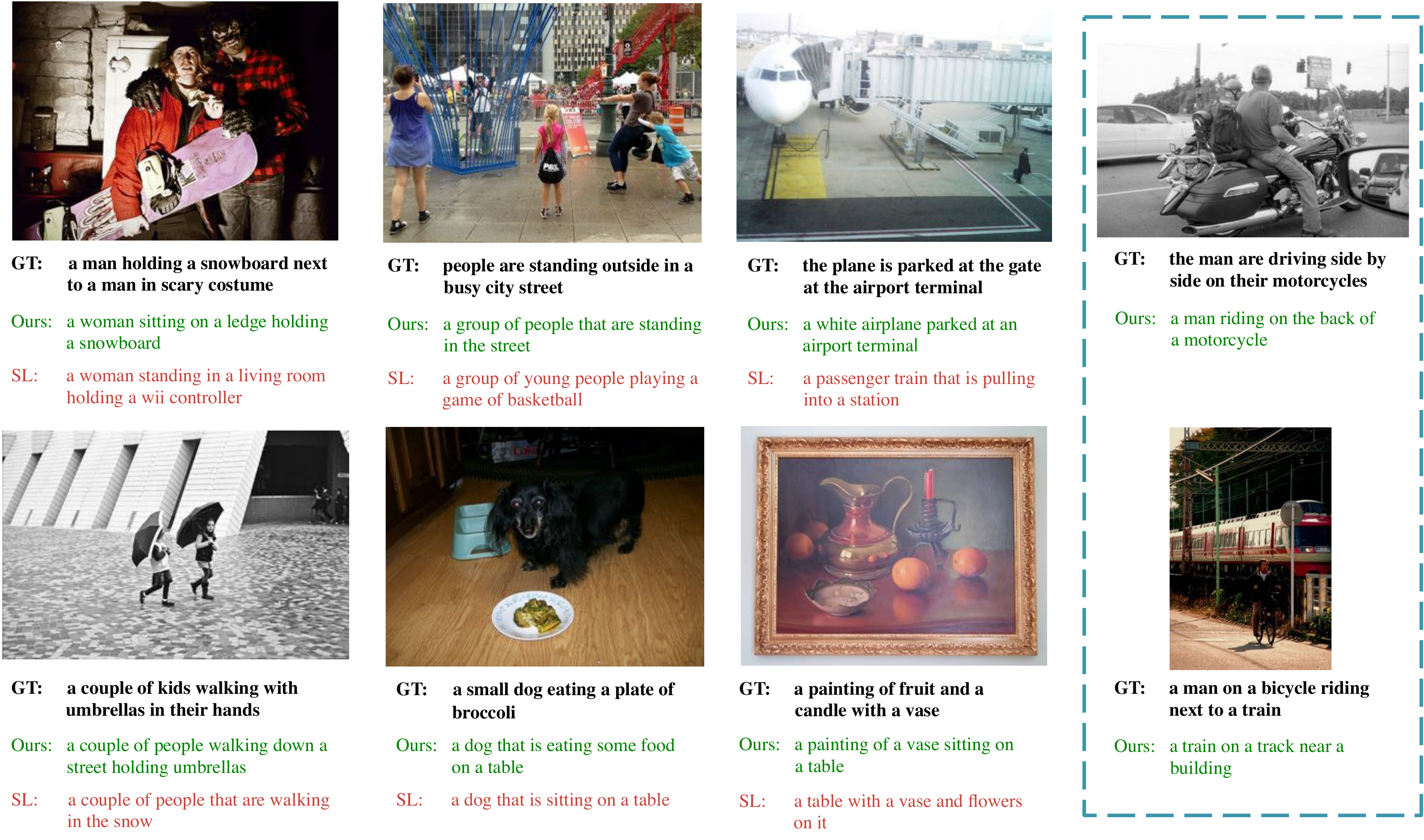}
  \vspace{-5pt}
\caption{ Qualitative results of our method and the supervised learning (SL) baseline. In the first three columns, our method generates better captions than SL. We show two failure cases in the last column. GT stands for ground truth caption.}
\label{FigExp}
\vspace{-5pt}
\end{figure*}

\subsection{How much each component contributes?}

In this section, we answer questions 1) 2) 3) above. As discussed in Section~\ref{training}, we train our policy and value networks in two steps: pretraining and then reinforcement learning. We name the initial policy network pretrained with supervised learning as ({\bf SL}). We name the initial value network pretrained with mean squared loss as ({\bf RawVN}). The SL model can be served as our baseline, which does not use value network or lookahead inference. To evaluate the impact of embedding, we incorporate SL with embedding as follows: in the last step of beam search of SL, when a beam of candidate captions are generated, we rank those candidates according to their embedding similarities with the test image other than their log-probabilities, and finally output the one with highest embedding score. This baseline is named as ({\bf SL-Embed}). To validate the contribution of lookahead inference and reinforcement learning, we construct a baseline that use SL and RawVN with the proposed lookahead inference, which is named as ({\bf SL-RawVN}). Finally our full model is named as ({\bf Full-model}). 

According to the results of those variants of our method shown in Table~\ref{table_compareVariants}, we can answer the questions 1)-3) above:
\begin{itemize} 
\vspace{-5pt}
\item[1.] Using embedding alone, {\bf SL-Embed} performs slightly better than the {\bf SL} baseline. However, the gap between {\bf SL-Embed} and {\bf Full-model} is very big. Therefore, we conclude that using embedding alone is not powerful. The proposed embedding-driven decision-making framework is the merit of our method.
\vspace{-7pt}
\item[2.] By using lookahead inference, {\bf SL-RawVN} is much better than the {\bf SL} baseline. This validates the importance of the proposed lookahead inference that utilizes both local and global guidance.
\vspace{-7pt}
\item[3.] After reinforcement learning, our {\bf Full-model} performs better than the {\bf SL-RawVN}. This validates the importance of using embedding-driven actor-critic learning for model training.
\vspace{-4pt}
\end{itemize}

We show some qualitative captioning results of our method and the SL baseline in Figure~\ref{FigExp}. GT stands for ground truth caption. In the first three columns, we compare our method and SL baseline. As we see, our method is better at recognizing key objects that are easily missed by SL, \emph{e.g.}, the \emph{snowboard} and \emph{umbrellas} in the first column images. Besides, our method can reduce the chance of generating incorrect word and accumulating errors, \emph{e.g.}, we generate the word \emph{eating} other than \emph{sitting} for the image in the lower second column. Moreover, thanks to the global guidance, our method is better at generating correct captions at global level, \emph{e.g.}, we can recognize the \emph{airplane} and \emph{painting} for the images in the third column. Finally, we show two failure cases of our method in the last column, in which cases we fail to understand some important visual contents that only take small portions of the images. This may be due to our policy network architecture. Adding more detailed visual modeling techniques such as detection and attention can alleviate such problem in the future.
\vspace{-5pt}

\begin{table*}
\centering
 \small
\begin{tabular}{| l | c c c c c c c | }
\hline
$\lambda$ &  Bleu{-}1 & Bleu{-}2 & Bleu{-}3 & Bleu{-}4 & METEOR & Rouge{-}L & CIDEr  \\
\hline
0 & 0.638 &0.471 & 0.34 & 0.247 & 0.233 & 0.501 & 0.8  \\
0.1 & 0.683 & 0.51 & 0.373 & 0.274 & 0.248 & 0.516 & 0.894 \\
0.2 & 0.701 & 0.527 & 0.389 & 0.288 & 0.248 & 0.521 & 0.922 \\
0.3 & 0.71 & 0.535 & 0.398 & 0.298 & \bf{0.251} & 0.524 & 0.934 \\
0.4 & \bf{0.713} & \bf{0.539} & \bf{0.403} & \bf{0.304} & 0.247 & \bf{0.525} & \bf{0.937} \\
0.5 & 0.71 & 0.538 & 0.402 & 0.304 & 0.246 & 0.524 & 0.934 \\
0.6 & 0.708 & 0.535 & 0.399 & 0.301 & 0.245 & 0.522 & 0.923 \\
0.7 & 0.704 & 0.531 & 0.395 & 0.297 & 0.243 & 0.52 & 0.912 \\
0.8 & 0.7 & 0.526 & 0.392 & 0.295 & 0.241 & 0.518 & 0.903 \\
0.9 & 0.698 & 0.524 & 0.389 & 0.293 & 0.24 & 0.516 & 0.895 \\
1 & 0.694 & 0.52 & 0.385 & 0.289 & 0.238 & 0.513 & 0.879 \\
\hline
\end{tabular}
\vspace{1pt}
\caption{Evaluation of hyperparameter $\lambda$'s impact on our method.}
\label{table_comparelambda}
\vspace{-4pt}
\end{table*}

\begin{table*}
\centering
 \small
\begin{tabular}{| l | c | c c c c c c c | }
\hline
Method & Beam size &  Bleu{-}1 & Bleu{-}2 & Bleu{-}3 & Bleu{-}4 & METEOR & Rouge{-}L & CIDEr  \\
\hline
\multirow{5}{*}{SL}
&5 & 0.696 & 0.522 & 0.388 & 0.29 & 0.238 & 0.513 & 0.876  \\
&10 & 0.692 & 0.519 & 0.384 & 0.289 &0.237 & 0.512 & 0.872 \\
&25 & 0.683 & 0.508 & 0.374 & 0.281 & 0.234 & 0.505 & 0.853 \\
&50 &  0.680 & 0.505 & 0.372 & 0.279 & 0.233 & 0.503& 0.850 \\
&100 & 0.679 & 0.504& 0.372& 0.279 & 0.233 & 0.503 & 0.849 \\
\hline
\multirow{5}{*}{Ours}
& 5 & 0.711 & 0.538 & 0.403 & 0.302 & 0.251 & 0.524 & 0.934 \\
&10 & {0.713} & {0.539} & {0.403} & {0.304} & {0.251} & {0.525} & {0.937} \\
&25 & 0.709 & 0.534 & 0.398 & 0.299 & 0.248 & 0.522 & 0.928 \\
&50 & 0.708 & 0.533 & 0.397 & 0.298 & 0.247 & 0.52 & 0.924 \\
&100 & 0.707 & 0.531 & 0.395 & 0.297 & 0.244 & 0.52 & 0.92 \\
\hline
\end{tabular}
\vspace{1pt}
\caption{Evaluation of different beam sizes' impact on SL baseline and our method.}
\label{table_compareBeamSize}
\vspace{-6pt}
\end{table*}

\subsection{Value network architecture analysis}
\label{netAnalysis}
\vspace{-2pt}

In this paper we propose a novel framework that involves value network, whose architecture is worth looking into. 
As in Figure~\ref{FigVN}, we use CNN$_v$ and RNN$_v$ to extract visual and semantic information from the raw image and sentence inputs. Since the hidden state in policy network at each time step is a representation of each state as well, a natural question is ``can we directly utilize the policy hidden state?''. To answer this question, we construct two variants of our value network: the first one, named as ({\bf hid-VN}), is comprised of a MLP$_v$ on top of the policy hidden state of RNN$_p$; the second variant, ({\bf hid-Im-VN}), is comprised of a MLP$_v$ on top of the concatenation of the policy hidden state of RNN$_p$ and the visual input $\bm{x}_0$ of policy RNN$_p$. The results are shown in Table~\ref{table_compareVariants}. As we see, both variants that utilize policy hidden state do not work well, comparing to our {\bf Full-model}. The problem of the policy hidden state is that it compresses and also loses lots of information. Thus, it is reasonable and better to train independent CNN, RNN for value network itself with raw image and sentence inputs, as in Figure~\ref{FigVN}.
\vspace{-3pt}

\subsection{Parameter sensitivity analysis}
\label{parAnalysis}
\vspace{-3pt}

There are two major hyperparameters in our method, $\lambda$ in Equation~\ref{lookaheadBS} and the beam size. In this section, we analyze their sensitivity to answer question 5) above. 

In Table~\ref{table_comparelambda}, we show the evaluation of $\lambda$'s impact on our method. As in Equation~\ref{lookaheadBS}, $\lambda$ is a hyperparameter combining policy and value networks in lookahead inference, $0\leq\lambda\leq1$. $\lambda=0$ means we only use value network to guide our lookahead inference; while $\lambda=1$ means we only use policy network, which is identical to beam search. 
As shown in Table~\ref{table_comparelambda}, the best performance is when $\lambda=0.4$. As $\lambda$ goes down from 0.4 to 0 or goes up from 0.4 to 1, overall the performance drops monotonically. This validates the importance of both networks; we should not emphasize too much on either network in lookahead inference. Besides, $\lambda=0$ performs much worse than $\lambda=1$. This is because policy network provides local guidance, which is very important in sequential prediction. Thus, in lookahead inference, it is too weak if we only use global guidance, \textit{i.e.} value network in our approach. 

In Table~\ref{table_compareBeamSize}, we provide the evaluation of different beam sizes' impact on SL baseline and our full model.  As discovered in previous work such as~\cite{karpathy_cvpr15}, the image captioning performance becomes worse as the beam size gets larger. We validate such discovery for existing encoder-decoder framework. As shown in the upper half of Table~\ref{table_compareBeamSize}, we test our SL baseline with 5 different beam sizes from 5 to 100. Note that SL is based on beam search, which follows the encoder-decoder framework as most existing approaches. As we see, the impact of beam size on SL is relatively big. 
It's mainly because that as we increase the beam size, bad word candidates are more likely to be drawn into the beam, since the confidence provided by the sequential word generator is only consider local information.

On the other hand, as shown in the lower part of Table~\ref{table_compareBeamSize}, our method is less sensitive to beam sizes. The performance variations between different beam sizes are fairly small. We argue that this is because of the proposed lookahead inference that considers both policy and value networks. With local and global guidances, our framework is more robust and stable to policy mistakes.


\vspace{-5pt}

\section{Conclusion}
\vspace{-3pt}
In this work, we present a novel decision-making framework for image captioning, which achieves state-of-the-art performance on standard benchmark. Different from previous encoder-decoder framework, our method utilizes a policy network and a value network to generate captions. The policy network serves as a local guidance and the value network serves as a global and lookahead guidance. To learn both networks, we use an actor-critic reinforcement learning approach with novel visual-semantic embedding rewards. We conduct detailed analyses on our framework to understand its merits and properties. Our future works include improving network architectures and investigating the reward design by considering other embedding measures.


{\footnotesize
\bibliographystyle{ieee}
\bibliography{egbib}
}

\end{document}